# Multilevel Threshold Based Gray Scale Image Segmentation using Cuckoo Search

Sourav Samanta[a*], Nilanjan Dey[b], Poulami Das[b], Suvojit Acharjee[c], Sheli Sinha Chaudhuri[c]

[a] Dept of CSE, Gobindapur Sephali Memorial Polytechnique, India
[b] Dept. of CSE, JIS College of Engineering, Kalyani, India.
[c] ETCE Dept., Jadavpur University, India

**Abstract**

Image Segmentation is a technique of partitioning the original image into some distinct classes. Many possible solutions may be available for segmenting an image into a certain number of classes, each one having different quality of segmentation. In our proposed method, multilevel thresholding technique has been used for image segmentation. A new approach of Cuckoo Search (CS) is used for selection of optimal threshold value. In other words, the algorithm is used to achieve the best solution from the initial random threshold values or solutions and to evaluate the quality of a solution correlation function is used. Finally, MSE and PSNR are measured to understand the segmentation quality.

## 1. Introduction

Image segmentation has a major importance in digital image processing. Image segmentation is actually the process of subdividing an image into its constituent regions or objects based on shape, colour, position, texture and homogeneity of image regions. In mathematical sense, the segmentation of image I, which is a set of pixels, is the partition of I into n disjoint sets $M_1, M_2,\ldots, M_n$, called segments or regions such that their union of all regions equals I. I= $M_1$ U $M_2$ U …. U $M_n$. There are many techniques for image segmentation. Thresholding is one of the widely used techniques for segmentation.

The segmented image obtained from thresholding has the advantage of smaller storage space, fast processing speed, and ease in manipulation, compared with a gray level image containing 256 levels. The aim of an effective segmentation is to replace the total image into a fewer number of levels without much loss in visual information. In literature many multi level methods are proposed [1, 2, 3]. This paper proposes a new method of multilevel optimal threshold based algorithm using cuckoo search [4, 5, 6].

In computer science, Meta heuristic designates a computational method that optimizes a problem by iteratively trying to improve a candidate solution with regard to a given measure of quality. Meta heuristics make few or no assumptions about the problem being optimized and can search very large spaces of candidate solutions. However, Meta heuristics do not guarantee an optimal solution ever found. Many Meta heuristics implement some form of stochastic optimization. The power of almost all modern Meta heuristics comes from the fact that they imitate the best feature in nature, especially biological systems evolved from natural selection over millions of years. Two important characteristics are selection of the best suited and adaptation to the environment. Numerically speaking, these are translated into two crucial characteristics of the modern Meta heuristics: intensification and diversification. Intensification intends to search around the current best solutions and select the best candidates or solutions, while diversification makes sure the algorithm can explore the search space efficiently.

------------
*Corresponding Author: Tel: 0-9732005674,
sourav.uit@gmail.com

In the paper *"Cuckoo search via Lévy flights"*, in: Proc. of World Congress on Nature & Biologically Inspired Computing (NaBIC 2009), December 2009, India. IEEE Publications, USA, pp. 210-214 (2009) Xin-She Yang and Suash Deb have introduced a new algorithm, called Cuckoo Search (CS), based on the interesting breeding behaviour such as brood parasitism of certain species of Cuckoos. Breeding behaviour of cuckoos and the characteristics of Lévy flights of some birds and fruit flies are introduced, and then the Cuckoo Search (CS) is formulated.

## 2. Methodology

### 2.1. Cuckoo Behaviour and Lévy Flights

#### 2.1.1. Cuckoo Breeding Behavior

In "Intra specific Brood Parasitism", Cuckoos lay their eggs in other birds' nests and leave, duping the hosts into rearing their offspring and freeing themselves in the process to lay more eggs instead of investing time and energy in parental care. In this way, they nick an important amount of time for reproduction which other species invest in rearing their young ones. In "Cooperative Breeding," Cuckoo's eggs generally hatch several days before the host's eggs, and the hosts 'chicks frequently starve in competition for food with the cuckoo chicks. This reduction in competition signifies that this behaviour increases survival of the parasite chick by [7] reducing the number of competing host chicks in the nest. In "Nest Takeover," the newly hatched cuckoo moves backwards up the side of the nest, pushing behind it an egg or young hosts' chick, finally hurling it down from the nest. It repeats this task until only it remains in the nest therefore taking over the food supply provided tirelessly by the foster parents. Therefore it shows how Cuckoos have responded to natural selection and adapted themselves in a way to increase the probability of survival of its species.

#### 2.1.2. Lévy Flights

Lévy flights are a class of random walks primarily devised by Paul [8, 9] Lévy in 1937 by generalizing Brownian motion to include non-Gaussian randomly distributed step sizes for the distance covered. Lévy Flights are random movements that can maximize the efficiency of resource searches in uncertain environments. It suggests that Lévy Flights provide a rigorous mathematical basis for separating out evolved, innate behaviours from environmental influences. It is also characterized by a considerable varied degree of step lengths and in some cases occurrence of extremely long jumps. In Optical science, Lévy flight can be defined as a term used to describe the motion of light. Sometimes, light travels in a random series of shorter and longer steps instead of travelling in a conventional Brownian diffusion. The shorter and longer steps together form a Lévy flight [10].

Most of the natural search processes are based on Lévy flights [11, 12, 13, 14]. After transferring to a new environment, bees perform Lévy flights to find the flowers in the area. The area covered by performing Levy flight is huger than the area covered by doing normal random search. Performing Levy Flight is also more informative than the other search techniques. Sharks follow random Brownian motion at the time of searching food; nevertheless, if they become unsuccessful, they exhibit Lévy flight behaviour, mixing short random movements with long trajectories.

Cuckoo Search can be described in a very simple manner by using the following three idealized rules:

a. A cuckoo chooses a nest randomly to lay its egg there. Each cuckoo lays one egg at a time.
b. The nests containing high quality eggs will carry over to the next generations.
c. The number of available host nests is fixed, and the egg laid by a cuckoo is discovered by the host bird with a probability p(0, 1).
   In this case, the host bird can either throw the egg away or discard the nest, and build a completely new nest. For simplicity, this last assumption can be approximated by the fraction of p with n nests that can be replaced by new nests (with new random solutions).

For a maximization problem, the quality or fitness of a solution can simply be proportional to the value of the objective function. Other forms of fitness can be defined in the similar way to the fitness function in genetic algorithms where the approach "better chromosome (solution) survives" is used. Here, to make the matter simple to understand, it has been assumed that the each egg in a nest represents a solution. A new solution is represented by a cuckoo egg. The aim is to use the new and potentially better solutions (cuckoo eggs) to replace a not-so-good solution in the nests. Of course, this algorithm can be extended to the more complicated case where each nest has multiple eggs representing a set of solutions.

Based on the above three rules, the basic steps of the Cuckoo Search (CS) can be summarized as the pseudo code shown below:

To generate a new solutions x (t+1), a cuckoo i is selected by Lévy flight search.

$$x_i^{t+1} = x_i^t + \alpha \oplus \text{Lévy}(\lambda)$$

Lévy flight is performed where α > 0. It is the step size which should be related to the scales of the problem of interests. In most cases, α = 1 is used. The above equation is an essential stochastic equation for random walk. In general, a random walk is a Markov chain whose next status/location only depends on the current location and the transition probability. The product means entry wise multiplications. This entry wise product is similar to those used in PSO. In our proposed work, the random walk via Lévy flight is used for its more efficient nature in exploring the search space. Its step length is much longer in the long run. The Lévy flight essentially provides a random walk while the random step length is drawn from a Lévy distribution.

$$\text{Lévy} \sim u = t^{-\lambda} \quad (1 < \lambda \leq 3)$$

which has an infinite variance with an infinite mean. Here the steps essentially form a random walk process with a power-law step-length distribution with a heavy tail.

Lévy walk is capable of generating a number of new solutions over the best solution obtained so far. Thus it speeds up the local search. To ensure that the system will not be trapped in a local optimum, a substantial fraction of new solutions having locations far enough from the current best solutions must be generated.

*2.1.2. Cuckoo Search via Lévy Flights*

*Begin*
Objective function f(x), x = ($x_1$, ..., $x_d$)T. Generate initial population of n host nests $x_i$ (i = 1, 2, ...,n)
  *While* (t <MaxGeneration) or (stop criterion)
    Get a cuckoo randomly by Lévy flights evaluate its quality /fitness Fi Choose a nest among
      n (say, j) randomly
      If (Fi > Fj), replace j by the new solution;
    End
      A fraction (pa) of worse nests are abandoned and new ones are built; Keep the best solutions
      (or nests with quality solutions); Rank the solutions and find the current best
  *End while*
  Postprocess results and visualization
*End*

## 2.2. Multilevel Image Segmentation based on Threshold

Image segmentation means dividing an image into non-overlapping regions that matches the real world objects. Complete segmentation divides an image $R$ into the finite number $S$ of regions $R_1, \ldots, R_S$

$$R = \bigcup_{i=1}^{s} R_i \quad \text{and} \quad R_i \cap R_j = \phi \quad \text{where} \quad i \neq j$$

Amongst the different segmentation [15, 16] methods, thresholding has drawn attention because of its simple nature. Thresholding techniques can be divided into bi-level and multi-level category, depending on the number of image segments. In bi-level thresholding, image is segmented into two different regions. The pixels with gray values greater than a certain value T are classified as object pixels, and the others with gray values lesser than T are classified as background pixels.

## 3. Proposed Method

Gray scale image contains pixel value within 0 to 255. In this approach this gray scale image generates initial solution of length equal to the number of levels. Size of this solution is similar to cuckoo's nest and each pixel value is considered as cuckoo's egg. For x no. of level there are $^{256}C_x$ no. of available possible candidate solutions. If gray scale image is segmented into four, then possible solutions are:

$$^{256}_{4}C = 174792640$$

For every one solution it is possible to get one segmented image. Amongst all the segmented images one of them is nearer to the original image. These similarities between original image and segmented image are measured according to the correlation value as given in equation 1. Here cuckoo search is applied to find the best candidate solution which gives the best correlation value.

The proposed algorithm is discussed below.

*Fitness Function:* This paper uses the following correlation as given in Eq-1 as fitness function for segmentation.
I and S represent original and segmented image respectively.
$\bar{I}$ and $\bar{S}$ are mean of original and segmented image. $\rho$ is the correlation value which measures the similarities between the segmented and original image. Its value is fractional value between 0 and 1.
$\rho = 1$ indicates the two images are same. Otherwise greater value of $\rho$ indicates there is very less change in original and segmented image.

$$\rho = \frac{\frac{1}{n^2}\sum_{i=1}^{n}\sum_{j=1}^{n}(I_{ij} - \bar{I})(S_{ij} - \bar{S})}{\sqrt{\frac{1}{n^2}\sum_{i=1}^{n}\sum_{j=1}^{n}(I_{ij} - \bar{I})^2}\sqrt{\frac{1}{n^2}\sum_{i=1}^{n}\sum_{j=1}^{n}(S_{ij} - \bar{S})^2}} \quad \ldots\ldots\ldots (1)$$

**Algorithm:**

Input: Original Image = I, No of Level = x
Begin
   Generate 20 initial solutions where each solution (nest) contains x no of eggs.
   Value of eggs will be a randomly generated number within 0 to 255.
  While (t < GENERATION) // stop criterion
   {
    Generate each cuckoo from egg via Lévy Flight.
    Segment the original image by individual cuckoo nest (i.e. candidate solution).
    Find the correlation of each segmented image with the original image I.
    Rank the correlation value and choose the CURRENT_BEST
    Correlation (fittest) value
      if ( CURRENT_BEST correlation > BEST  correlation)
     BEST=CURRENT_BEST
     store the corresponding solution into the BEST_NEST
    End
    Some nests are deleted with probability pa by host bird and created new value.
   }
  End While
      t=t+1
  Print the BEST correlation and  BEST_NEST   // this is the best solution that gives the
  Segment the image I by BEST_NEST              // best segmentation result with highest
                                                     //correlation value
End

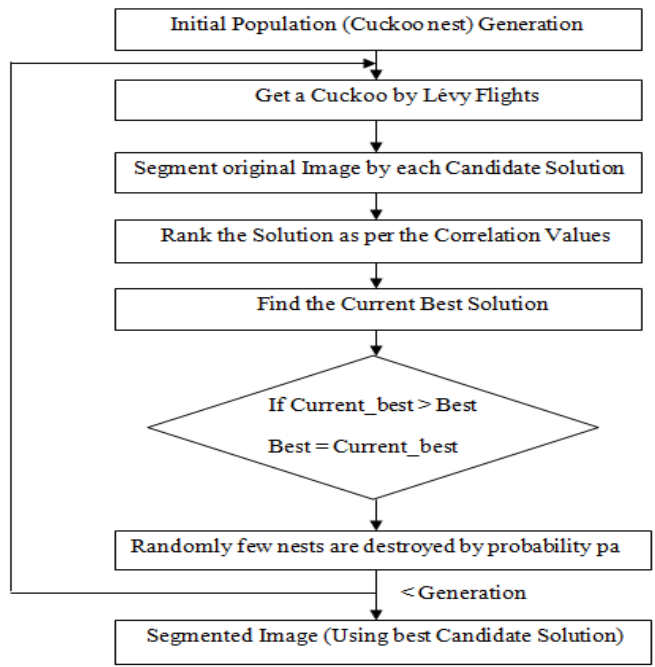

Figure 1: Block Diagram of the Algorithm of Multilevel Image Segmentation Based on Threshold using Cuckoo Search

## 4. Result and Discussions

MATLAB 7.0.1 Software is extensively used for the study of Image segmentation. Concerned images obtained in the result are shown in Figure 2 through 5.

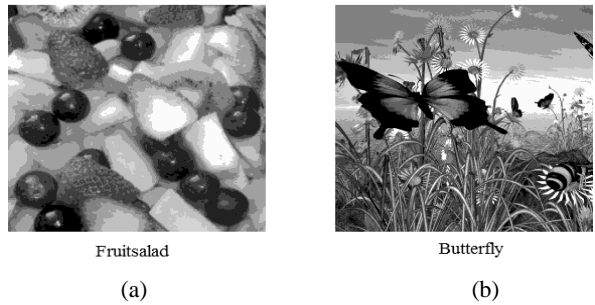

Figure 2: (a) Original Fruitsalad Image, (b) Original Butterfly Image

The proposed algorithm is applied on the two images as given in Fig 2. These two images have long range and well distributed histogram. Each image is segmented with level 2, 4,6 and 8 respectively. For this experimentation cuckoo size is 20. Iteration is 50. Probability of nest discovered by host bird is 0.25. Result of segmentation is shown in Fig 3.

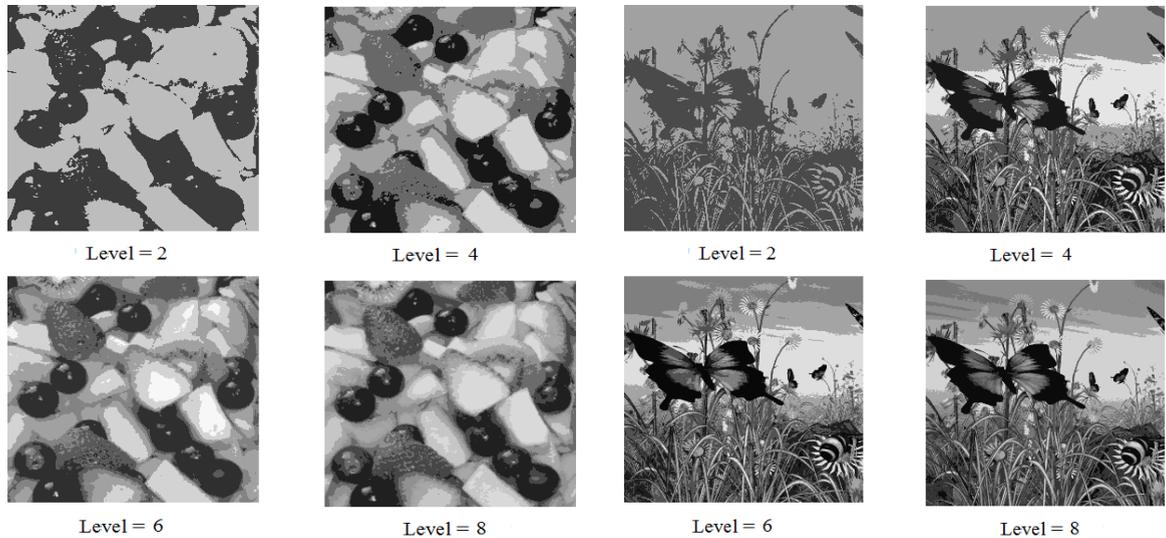

Figure 3: (a) Level 2, Level 4, Level 6, Level 8 Segmented Fruitsalad Image, (b) Level 2, Level 3, Level 4, Level 8 Segmented Butterfly Image

Table 1

| Level | Correlation Value | |
|---|---|---|
| | Fruit salad | Butterfly |
| 02 | 0.862365678098395 | 0.837546411216321 |
| 04 | 0.960600224380122 | 0.958614685667217 |
| 06 | 0.979351284121812 | 0.978339132027312 |
| 08 | 0.987663274469147 | 0.986633861241287 |
| 10 | 0.992039321158848 | 0.990339838073647 |
| 20 | 0.996863284560225 | 0.996583117993407 |
| 40 | 0.998995482363943 | 0.998890439024449 |
| 50 | 0.999322671778229 | 0.99931680876461 |

Peak Signals to Noise Ratio (PSNR):

$$PSNR = 20*\log_{10}\left(255/\sqrt{MSE}\right) \quad \ldots(2)$$

where, MSE is the cumulative squared error between the segmented image and the original image. PSNR is a measure of the peak error. The mathematical formula of MSE is

$$MSE = \frac{1}{MN}\sum_{y=1}^{M}\sum_{x=1}^{N}[I(x,y)-S(x,y)]^2 \quad \ldots(3)$$

where, I(x,y) is the original image, S(x,y) is the segmented image and M, N are the dimensions of the images. A lower value for MSE means lesser error, and as seen from the inverse relation between the MSE and PSNR, this translates to a high value of PSNR. Logically, a higher value of PSNR is good because it means that the ratio of Signal to Noise is higher. Here, the 'signal' is the original image, and the 'noise' is the error in segmentation process. So if high PSNR is found, it indicates good quality segmentation. Fig. 5 shows the result of under different segmentation of two images.

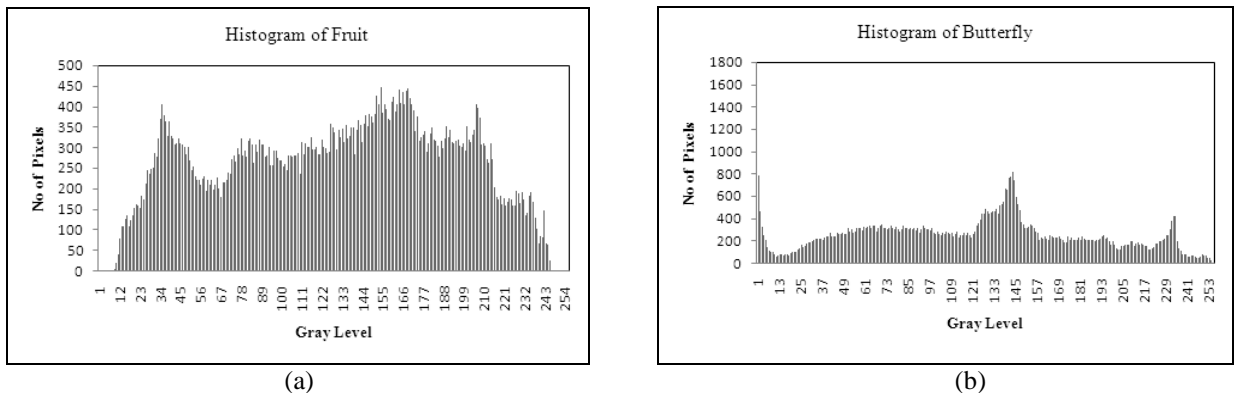

(a)            (b)

Figure 4: Histogram Analysis of Original Fruitsalad Image, (b) Histogram Analysis of Original Butterfly Image

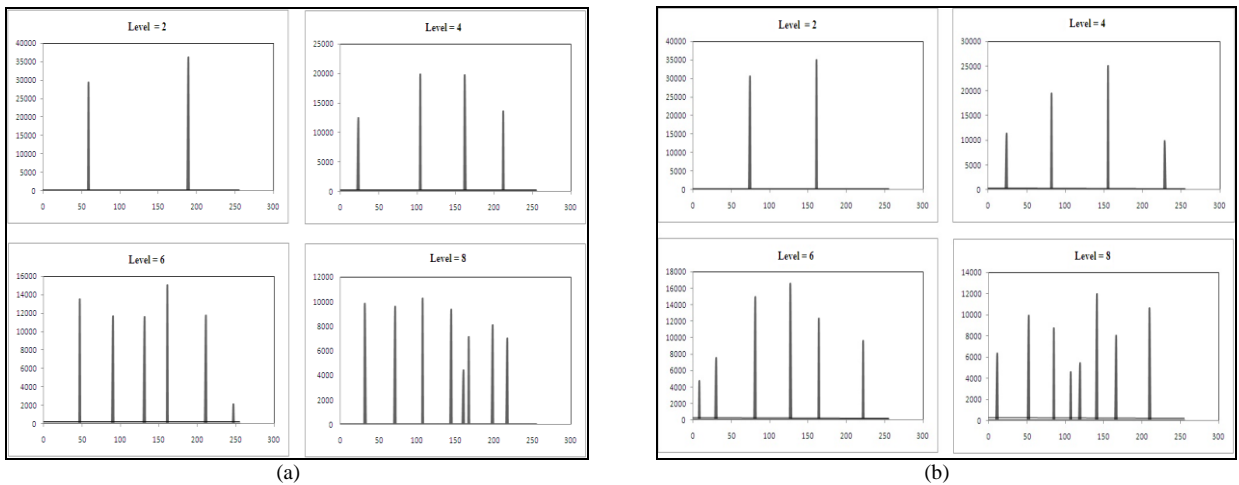

(a)            (b)

Figure 5: (a) Histogram Analysis of level 2, level 4, level 6, level 8 Segmented Fruitsalad Images, (b) Histogram Analysis of level 2, level 4, level 6, level 8 Segmented Butterfly Images

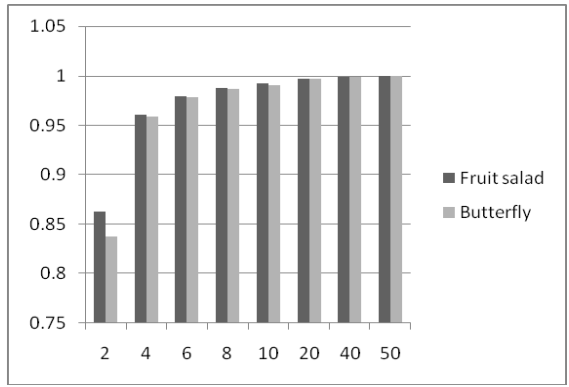 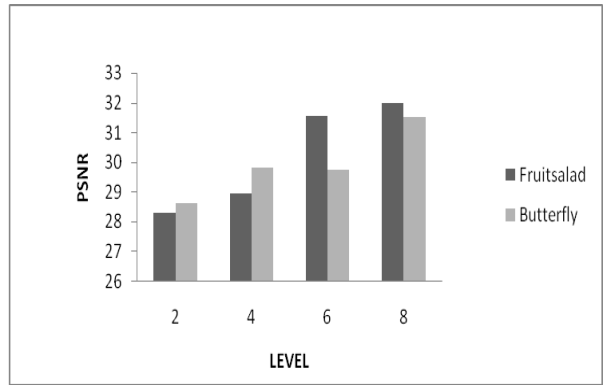

Figure 6: Correlation vs. Level         Figure 7: PSNR vs. Level

## 5. Conclusion

The robustness of segmentation increases by tuning the probability value of abandoned nest (pa), which can change the diversification of the search space. This proposed approach is comparatively faster than any other existing genetic algorithm based segmentation algorithms. Cuckoo search based segmentation can also be applicable in the cases of multi objective based image segmentation. This segmentation methodology can be widely applicable depending upon various problem domains.